\documentclass[letterpaper, 10 pt, conference]{ieeeconf} 
\usepackage{graphicx, amsmath, cleveref }
\usepackage{hyperref}
\usepackage{color}
\usepackage[noadjust]{cite}
\usepackage[normalem]{ulem}

\title{\LARGE \bf
Word2vec to behavior: morphology facilitates the grounding of language in machines.}

\author{
\authorblockN{David Matthews}
\authorblockA{University of Vermont}
\and
\authorblockN{Sam Kriegman}
\authorblockA{University of Vermont}
\and
\authorblockN{Collin Cappelle}
\authorblockA{University of Vermont}
\and
\authorblockN{Josh Bongard}
\authorblockA{University of Vermont}
}

\begin{document}

\teaser{
\centering
\includegraphics[width=\linewidth]{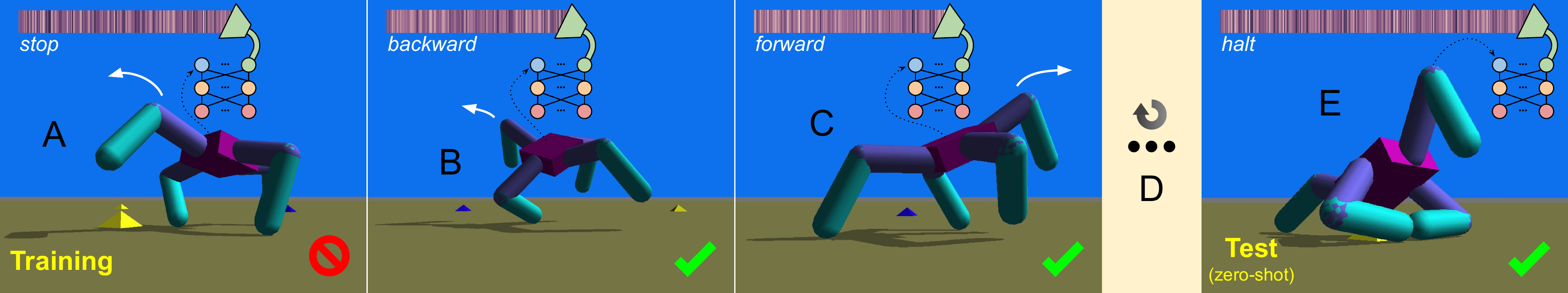}
\vspace{-1.5em}
\caption{\textbf{Overview of the method.}
\textbf{A:} The initial values of a robot control policy's hidden
layer is set by supplying  
the word2vec embedding associated
with a command such as `stop' to one neuron in the
input layer. The policy is then
downloaded on to a robot, and the sensor data
generated by its movement is supplied to the remainder
of the input layer (dotted arrow), further
altering the hidden- and motor layers.
After evaluation, the robot's behavior is scored against
an objective function paired with the command, 
such as one that penalizes motion.
The same policy is then evaluated 
four more times (two of which are shown in \textbf{B} and \textbf{C})
against four other commands and objective functions.
Policies are trained to maximize mean performance
against all five of these functions (\textbf{D}). After training, the best policy
is supplied with a sixth, previously-unheard synonym of
`stop', and its behavior is scored against the `stop' objective function (\textbf{E})
(\href{https://youtu.be/MYegNCJ5bWU}{\textcolor{blue}{\textbf{\texttt{youtu.be/MYegNCJ5bWU}}}}).
}
\label{fig:teaser}
\vspace{-2em}
}
\maketitle
\thispagestyle{empty}

\begin{abstract}

Enabling machines to respond appropriately to natural language
commands could greatly expand the number of people to whom they
could be of service.
Recently, advances in neural network-trained word embeddings
have empowered non-embodied text-processing
algorithms, and suggest they could be of similar utility for
embodied machines.
Here we introduce a method that does so by training robots
to act similarly to semantically-similar word2vec encoded commands. 
We show that this enables them to act
appropriately, after training, to previously-unheard commands. Finally,
we show that inducing such an alignment between motoric and
linguistic similarities can be facilitated or hindered by
the mechanical structure of the robot. This points to
future, large scale methods that find and exploit
relationships between action, language, and robot structure.

\end{abstract}

\section{Introduction}
\label{sectIntro}

Using natural language to interact with machines has long been a goal
in AI research. Recently, word embeddings such as word2vec have yielded
significant advances in this direction \cite{mikolov2013distributed, mikolov2013efficient}.
These embeddings generate vector spaces which accurately preserve semantic
relationships between words, and can
then be used to address 
text classification \cite{lilleberg2015support}, 
sentiment analysis \cite{tang2014learning, dickinson2015sentiment},
and other natural language-based problems.

However, these approaches tend to disregard the role that action and
the body of an agent may play in generating and understanding
natural language. 
The link between action and natural language has
long been hypothesized in 
cognitive science \cite{clark2006language}
and 
linguistics \cite{lakoff2008metaphors}. 
However, it was only recently that
neuroscience studies have provided data suggesting that such a
link exists \cite{gallese2005brain}. 
For instance, Pulvem\"{u}ller {\it et al.} have shown that
if stories are read to immobile subjects being scanned by fMRI,
their motor and sensor cortices exhibit heightened activity
\cite{pulvermuller2010active}.

In robotics, a long literature in helping robots ground
language in action exists \cite{selfridge1986natural}.
For example,
Steels \textit{et al.} reported a series of experiments in which robots collectively
construct their own syntax and grammar \cite{steels2003evolving}, 
while Schulz \textit{et al.} report on robots
that construct a language to describe 
spatial \cite{schulz2011lingodroids} 
and temporal \cite{heath2012lingodroids}
concepts. Matuszek \textit{et al.} \cite{matuszek2013learning} 
trained a parser on pairs of English
commands and  corresponding control language expressions.

The word embedding approach is attractive as a data-driven,
rather than hypothesis-driven, method for enabling machines
to link natural language and action. Indeed, recent such
attempts have been reported. For example a visual word2vec
corpus has been trained which captures semantic relationships
between images rather than text
\cite{kottur2016visual}, 
and sound-word2vec similarly
discovers semantic structure among the sounds associated
with words \cite{vijayakumar2017sound}. Jang \textit{et al.} 
demonstrated
``grasp2vec'': a method for enabling robots to autonomously 
learn object-centric representations that enable recognition
and grasping of objects without recourse to a pre-defined
feature set \cite{jang2018grasp2vec}.

However, none of these methods attempt to align embodied embeddings with word embeddings.
In order to realize robots that can respond appropriately to previously unheard natural language,
it would be useful if the robot's learned sensorimotor
structure mapped on to natural language semantic structure, and vice versa.

If so, robots should act similarly 
and appropriately
when they hear two similar words, even if they have not previously heard one of those words.

We demonstrate a method that forges such
an alignment here. Briefly, we assign a unique objective function
to sets of similar action words, and then train robots to maximize
these functions while they ``hear'' the embeddings associated with
those words, along with their own sensor data. Evidence that
robots can successfully learn to align human language semantic
structure with the structure of their own felt experience is demonstrated
by the fact that robots so trained act appropriately when issued
previously unheard words.

Finally, we have found that the mechanical design of the robot
can facilitate or obstruct the training algorithm's ability to
forge this alignment: we found the method performed worse or better
for robots with different body plans. This adds to a growing
body of work that demonstrates that an appropriate 
robot body plan choice 
can facilitate other aspects of behavior generation in robots 
\cite{bongard2011morphological, bongard2015evolving,mahoor2017morphology,kriegman2018morphological,kriegman2019automated}.

\section{Methods}
\label{sectMethods}

\subsection{The task.}
\label{sec:task}

Robots were optimized in simulation using
\textit{Pyrosim}\footnote{Pyrosim is a python interface for building robots (and their neural controllers) in Open Dyanamics Engine: \href{https://github.com/mec-lab/pyrosim}{\color{blue}\textbf{\texttt{github.com/mec-lab/pyrosim}}}}
to perform three behaviors (move forward, move backward, and stop movement) according to the 
embeddings
of six different input commands: `forward', `backward', `stop', `cease', `suspend', and `halt'.
Prior to optimization, one of the last four commands (i.e., one of the synonyms of `stop') is randomly selected for testing and held out of the training set.
Robots are optimized according to their performance summed across all five training commands.

The performance of a robot under the `forward' and `backward' commands was measured by their respective displacement in the positive and negative $x$-axis of the simulator, at the end of an evaluation period of 500 time steps  (with step size 0.05; i.e., 25 seconds of behavior).
For the remaining commands, performance was proportional to the negative euclidean distance from the origin (the robot's starting point) at the end of simulation, thus rewarding robots that move less.

Because robots were tested under an unheard synonym of `stop',
test error was measured as
the final displacement of the robot.

\subsection{The controller.}
\label{sec:controller}

\begin{figure}[t]
\centering
\includegraphics[width=0.85\linewidth]{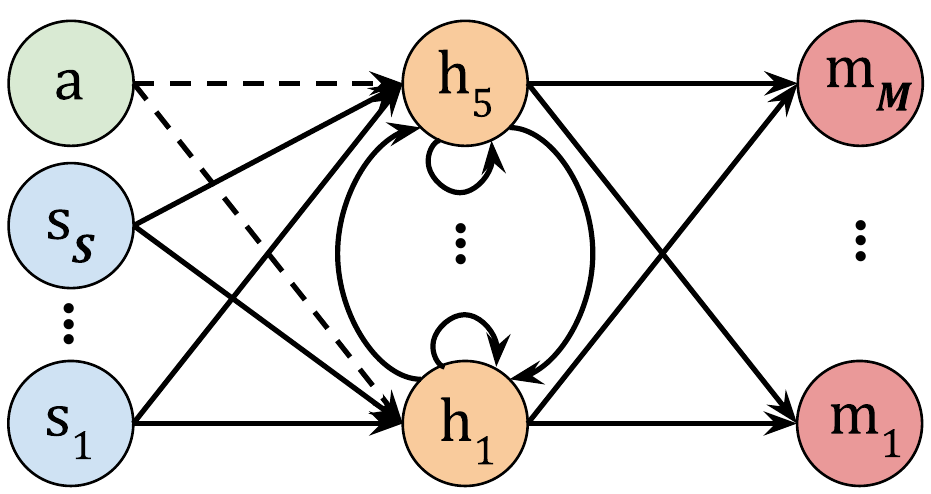}
\caption{
\textbf{Neural network architecture.} 
Controllers here contain four neuron types: auditory (green; $a$), sensor (blue; $s_i$), hidden (orange; $h_i$), and motor (red; $m_i$) neurons.
Prior to behavior, the auditory neuron is attached to the recurrent, hidden layer, and a command vector is fed in serially along the auditory neuron's synapses into the hidden layer.
The synapses connecting the auditory neuron to the hidden neurons are then detached, and the robot is sent to the simulator with only the sensor, hidden, and motor neurons. 
Robots here have between 0 and 4 sensors, and between 1 and 8 motors.
}
\label{fig:network}
\vspace{-1em}
\end{figure}

The robots are controlled by recurrent neural networks with three layers: a sensor layer, which is fully connected to a
self- and recurrently-connected
hidden layer consisting of five neurons, which are fully connected to a motor layer (Fig.~\ref{fig:network}). 
The number of motor neurons and sensor neurons vary with the morphology of each robot.

The sensor layer includes an \textbf{auditory neuron} that initializes the controller's hidden state as follows.
Before sending a robot to the simulator for evaluation, the target command vector is fed serially through the auditory neuron and into the recurrent hidden layer, one element after another, each time updating the hidden neurons' values. 

After initializing the hidden neurons, their incoming synapses from the auditory input neuron were removed, and the sensor and motor neurons were attached.
The robot was then sent to the simulator with its initialized network
(Fig.~\ref{fig:W2VInitHiddenNeurons}).

\subsection{The robots.}
\label{sec:robot}

\subsubsection{The quadruped}
The quadruped (Fig.~\ref{fig:teaser}) consists of a rectangular abdomen, attached to which are four legs, each composed of an upper and lower cylindrical object. 
The knee and the hip joint of each leg contain a 1-DOF rotational hinge joint which can flex inward or extend outward by up to 45 degrees away from its initial angle (Fig.~\ref{fig:robots}C). 
Inside each lower leg is a touch sensor neuron, which at every time step detects contact with the floor: 
its
value is either $0$ (no contact) or $+1$ (contact).

\subsubsection{The minimal robot}

The minimal robot (Fig.~\ref{fig:robots}A) consists simply of two cylinders joined end-to-end by a rotational hinge joint.
Like a single leg of the quadruped, minimal robots have a single degree-of-freedom hinge joint.
However, unlike the quadruped's legs, the minimal robot has two touch sensors (one in each cylinder) as well as a proprioceptive sensor that measures the angle of its joint.

\subsubsection{The spherical robots}

\begin{figure}[h]
\centering
\includegraphics[width=\linewidth]{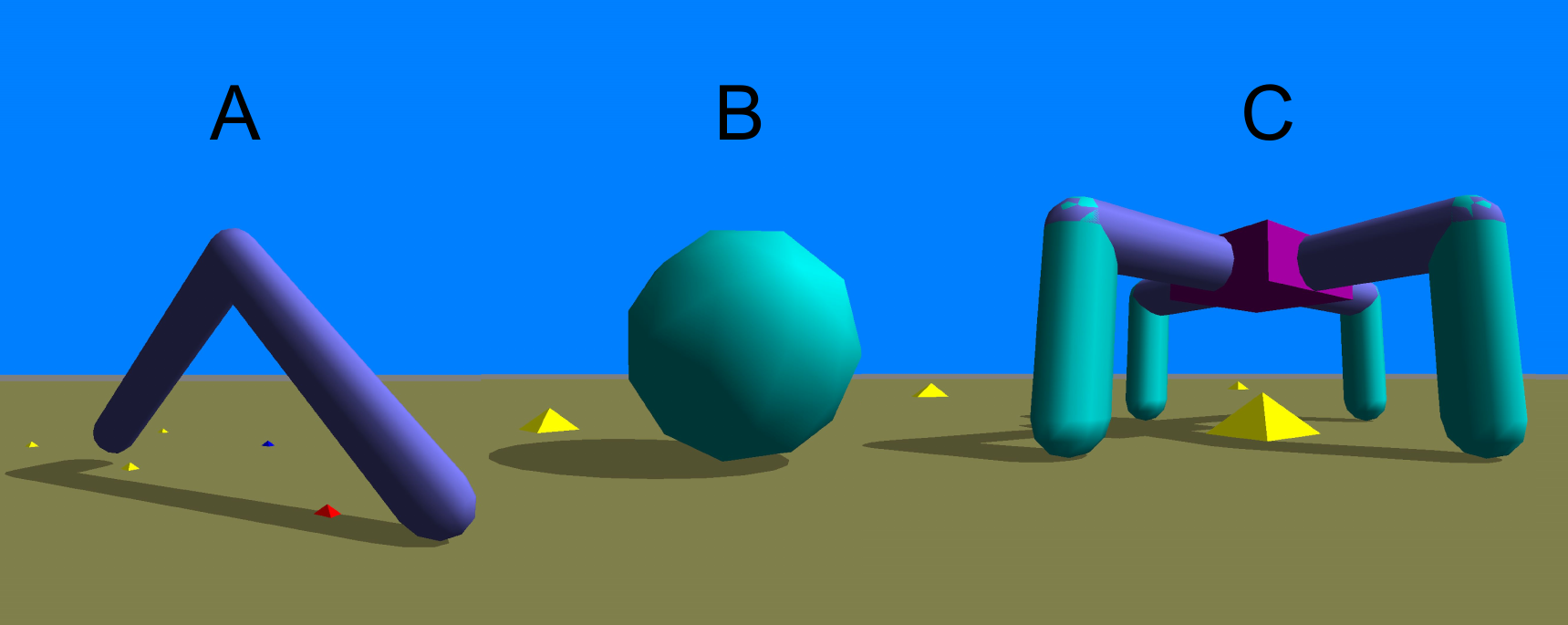}
\caption{\textbf{Three robots:}
Minimal (\textbf{A}), spherical (\textbf{B}), and quadrupedal (\textbf{C}).
}
\label{fig:robots}
\end{figure}

The spherical robots (Fig.~\ref{fig:robots}B) consist of a pendulum attached to a sphere's top interior wall. 
Some spherical robots have a pendulum which can only swing through the $xz$ plane (where $z$ is the vertical axis). 
These are referred to as 1DOF spherical robots. 
Other spherical robots have two orthogonal joints rotating in both the $xz$ and $yz$ planes. 
This version is referred to as a 2DOF spherical robot. 
Some spherical robots have proprioception (denoted as ``spherical robots with sensors'') and others do not (denoted as
``without sensors'').

\begin{table}[ht]
    \centering
    \label{tbl:CosineSimilarity}
    \caption{Cosine similarity between different commands in the \textbf{word2vec embedding.}}
    \vspace{-4pt}
    \begin{tabular}{|c|c|c|c|c|c|c|} \hline
        & forward & backward & halt & stop & suspend & cease \\ \hline
        forward &1& 0.42 & 0.15 & 0.13 & 0.11 & 0.02 \\ \hline
        backward &&1& 0.17 & 0.17 & 0.09 &-0.01\\ \hline
        halt &&&1& 0.61 & 0.63 & 0.56 \\ \hline
        stop &&&&1& 0.38 & 0.50 \\ \hline
        suspend &&&&&1& 0.57 \\ \hline
        cease &&&&&&1 \\ \hline
    \end{tabular}
\end{table}

\begin{table}[ht]
\centering
    \label{tbl:CosineSimilarityPermuted}
    \caption{Sample cosine similarity between commands in the word2vec embedding, after \textbf{randomly-permuting} the vector.}
    \vspace{-4pt}
    \begin{tabular}{|c|c|c|c|c|c|c|} \hline
        & forward & backward & halt & stop & suspend & cease \\ \hline
        forward &1& 0.03 & 0.06 & 0.05 & 0.04 &-0.05 \\ \hline
        backward &&1& 0.09 &-0.01 &-0.00 &-0.00\\ \hline
        halt &&&1&-0.02 & 0.11 &-0.05\\ \hline
        stop &&&&1& 0.04 &-0.01 \\ \hline
        suspend &&&&&1&-0.04 \\ \hline
        cease &&&&&&1 \\ \hline
    \end{tabular}
\end{table}

\begin{table}[!ht]
\centering
    \label{tbl:Fitness}
    \caption{Median (and standard deviation) \textbf{training performance,} with Holm-Bonferroni correction for 56 comparisons.}
    \vspace{-4pt}
    \begin{tabular}{|l|c|c|c|} \hline
        Robot                       & Control           & Word2vec          & P-Value           \\ \hline
        Quadruped                   & 1.84  (0.07)    & 1.90  (0.49)    & $p > 0.05$ \\ \hline
        Minimal                     & 5.30  (0.06)    & 5.43  (0.07)    & $p > 0.05$ \\ \hline
        1DOF Spherical & & & \\ with sensors    & 10.74 (0.08)    & 11.22 (0.08)    & $ \mathbf{p <0.05}$\\ \hline
        1DOF Spherical & & & \\ without sensors & 11.48 (0.08)    & 11.20 (0.09)    & $p > 0.05$ \\ \hline
        2DOF Spherical & & & \\ with sensors    & 10.55 (0.08)    & 10.84 (0.08)    & $p > 0.05$  \\ \hline
        2DOF Spherical & & & \\ without sensors & 10.61 (0.09)    & 10.33 (0.07)    & $p > 0.05$ \\ \hline
    \end{tabular}
    \vspace{-1em}
\end{table}

\subsection{The optimization algorithm.}
\label{sec:optimization}

Controllers were optimized using a standard evolutionary algorithm: AFPO (Age-Fitness Pareto Optimization; \cite{schmidt2011age}).
AFPO is a multi-objective optimization method that trains populations
of candidate solutions to maximize the objective
function for the desired behavior, while simultaneously minimizing `age', a variable which roughly corresponds to the amount of search time spent in a particular area of design space. 
This latter objective aids in the prevention of premature convergence.

Each independent evolutionary run started with a different random seed, and consisted of a population of 50 robots, optimized for 6000 generations. 
At each generation, modified copies are made of each robot in the population by randomly selecting a single synapse and perturbing it according to the normal distribution with a mean of the current synapse weight value, and standard deviation of the absolute value of the current synapse weight.

\subsection{The experimental treatment.}
\label{sec:experimental}

Prior to optimization, the vectors corresponding to each command were obtained from the word2vec embedding.\footnote{\href{https://code.google.com/archive/p/word2vec/}{\textcolor{blue}{\textbf{\texttt{code.google.com/archive/p/word2vec}}}}}
During optimization, a command vector was uploaded to the robot (as described in \S\ref{sec:controller}); then, the robot behaved and was assigned a performance score (as described in \mbox{\S\ref{sec:task}}).
The cosine similarities between pairs of the command vectors are presented in Table~I.

\subsection{The control treatment.}
\label{sec:control}

There is a possibility for overfitting in our method due to the unbalanced
nature of the training set. Since the majority of the training commands 
(three out of five) require the robot to remain stationary, control policies
could evolve that keep the robot immobile by default, yet memorize a movement
response for the `forward' command and another for the `backward' command. 
In this way, even if we observe that the robot stays immobile when presented with the held-out, fourth
`stop' synonym, the control policy causing this behavior may have
ignored the latent structure in the command embeddings.

In order to assess whether such overfitting occurs, we use the following control.
At the beginning of each evolutionary run, the vectors corresponding to each command were obtained from the word embeddings vector space. Each vector was then randomly permuted so that the distribution of values in each new vector do not change, but their orderings do (Table~II). The resulting five permuted embeddings are held constant over the course of that evolutionary run.
If the optimization method tends to yield overfit control policies, they should similarly keep
the robot immobile when presented with the sixth, held-out `stop' synonym, regardless of the permutation.

If however the control policies exploit the latent structure in the embeddings and that structure is disrupted by permutation, we should expect to see the control treatment policies generate more movement 
in the robot, compared to the experimental treatment policies, when
both are presented with the held-out `stop'
synonym. In other words, the control treatment policies
should generalize worse than the experimental treatment policies 
when presented with the test command.

\subsection{The hypothesis tests and correction.}
\label{sec:stats}

For hypothesis testing, we use the
Mann-Whitney~U test \cite{mann1947test}, a rank-based test of whether one of two random variables is larger than the other. 

We make a total of 56 pairwise comparisons in this paper.
With each comparison, the
likelihood of incorrectly rejecting a null hypothesis (i.e., making a Type I error) increases.
Thus, 
to control the family-wise error rate (the probability of one or more false rejections of true hypotheses) 
we conservatively adjust the rejection criteria of each individual hypothesis test using the the Holm-Bonferroni (step-down) procedure \cite{holm1979simple}.

\begin{figure*}
\centering
\includegraphics[width=1.0\linewidth]{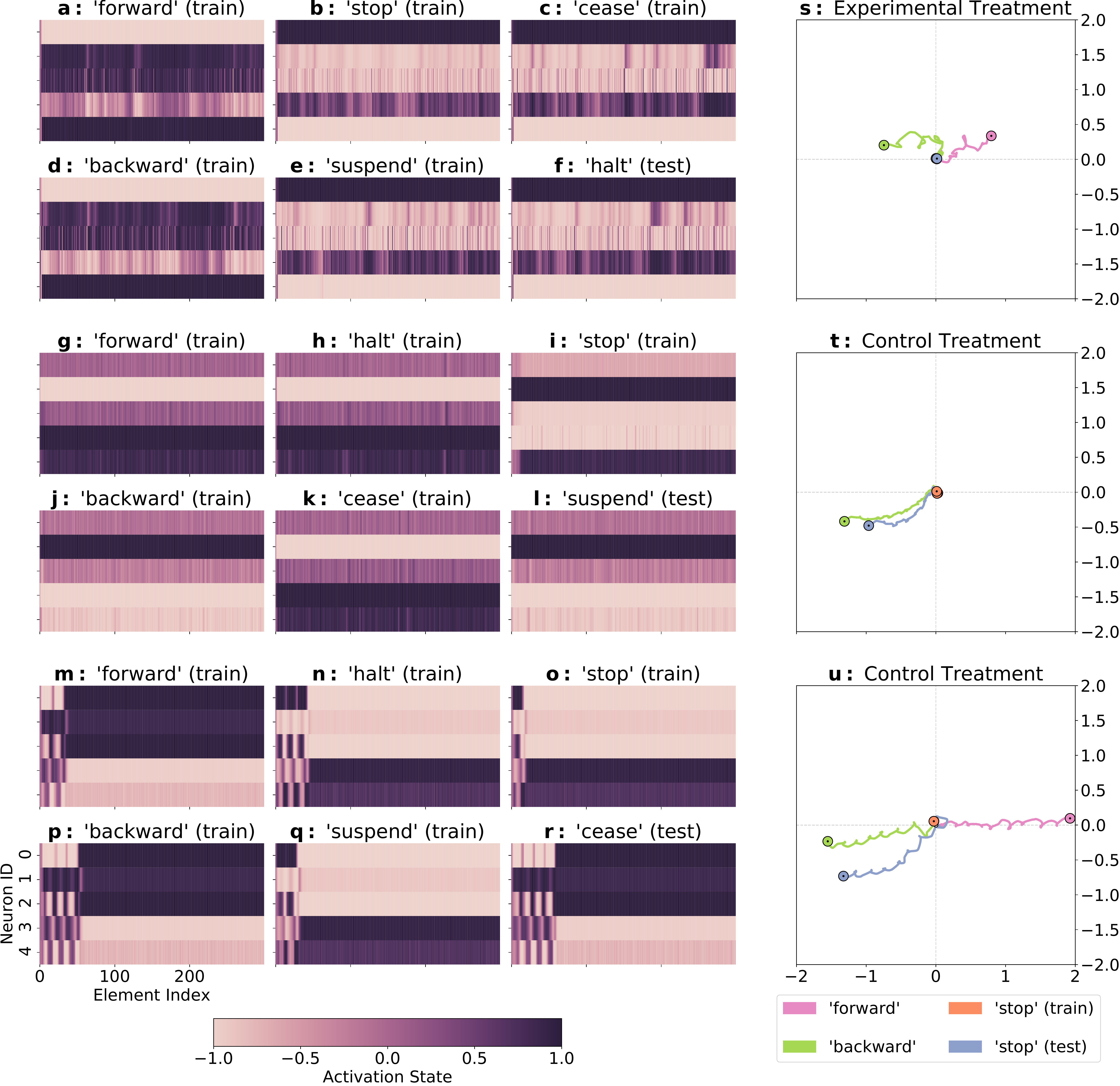}
\caption{\textbf{Initialization and behavior of three quadrupedal controllers.} 
The robot's five hidden neurons were initialized with the six different command vectors, prior to behavior, by feeding the vectors serially into the network (\textbf{a-f}).
After the hidden layer is fully initialized (the rightmost column of a-f), the robot behaves in closed-loop with feedback from its touch sensors (\textbf{s}; starting at the origin).
\textbf{s:} This controller successfully generated correct behavior for each of the word2vec training commands and also achieved low test error by suppressing movement when commanded to `halt' even though `halt' was held out of the training set.
On visual inspection of a-f, it seems as though neurons 0 and 4 encode motion, while the others (neurons 1-3) may encode direction.
\textbf{t:} When trained with randomly-permuted word2vec command vectors (see \S\ref{sec:control} for details), this controller failed to yield correct behavior during training:
the robot remains close to the origin despite being rewarded to move `forward' (\textbf{g}; pink trajectory in t).
During testing, when commanded to `suspend' (\textbf{l}; orange trajectory in t), the robot moved similarly to how it moved when commanded to go `backward' (\textbf{j}; green trajectory in t).
\textbf{u:} This controller was successfully trained to generate appropriate behavior with randomly-permuted word2vec command vectors, but failed to generalize, and thus behaved inappropriately when commanded to `cease': the robot moves as though commanded to go backward (\textbf{r} resembles \textbf{p}; the blue trajectory follows the green trajectory).
These counterfactuals (g-r; t and u) indicate that successful controllers in the experimental treatment utilized the word2vec embedding to achieve zero-shot generalization.
}
\label{fig:W2VInitHiddenNeurons}
\end{figure*}

\begin{figure*}
\centering
\includegraphics[width=0.95\linewidth]{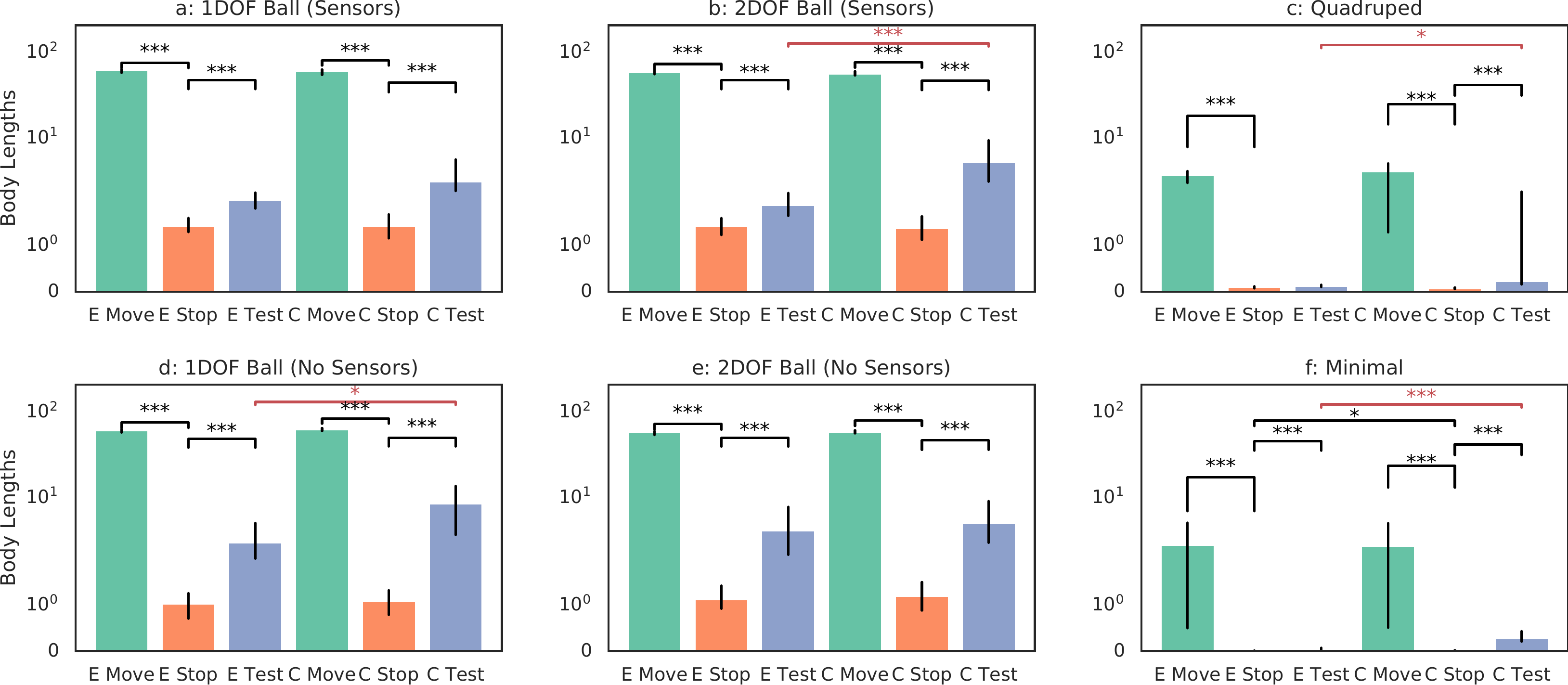}
\caption{\textbf{The effect of morphology on symbol grounding.}
    Median net displacement in body lengths of the run champions (the best robot from each optimization trial) grouped by morphology and command meaning: ``Move'' denotes the commands `forward' and `backward' during training, ``Stop'' denotes the training `stop' commands, and ``Test'' denotes the test `stop' command (i.e., test error).
    The experimental treatment (word2vec vectors) is denoted by ``E'', and the control treatment (randomly-permuted word2vec vectors) is denoted by ``C''. 
    For pairwise comparisons, we used the Mann-Whitney U test with Holm-Bonferroni correction for 56 comparisons (see \S\ref{sec:stats}). 
    The red significance brackets specifically denote a difference between the experimental and control treatments in terms of zero-shot generalization.
    The presence of red brackets for some morphologies, but not for others, indicates that only certain morphologies were able to align their sensorimotor structure with the word2vec embedding.
    Error bars denote 95\% bootstrapped confidence intervals of the median. 
    A single asterisk denotes significance at the 0.01 level, two denote the 0.001 level, and three the 0.0001 level. 
    }
\label{fig:testError}
\vspace{-1em}
\end{figure*}

\section{Results}
\label{sectResults}

Twelve hundred independent evolutionary trials were performed in total: 100 for each of the experimental and control treatments, for each of the six robot morphologies (Table~III).

The 1200 run champions|the best robot from each trial|are extracted in order to test for statistical differences between the treatments, commands and morphologies.

Fig.~\ref{fig:W2VInitHiddenNeurons} traces the behaviors of three exemplar run champions that are representative of typical behaviors found by the optimizer in the control and experimental treatments.
Under the control treatment, the optimizer yielded specialized robots that were unsuccessful at one or more of the training behaviors and failed to ``understand'' the meaning of the unheard synonym: they have high test error.
Under the experimental treatment, the optimizer yielded robots with correct behavior on all three training commands and that understood the meaning of the unheard synonym: they have low test error.

Fig.~\ref{fig:testError} compares the average displacement of the run champions under the different experimental conditions tested here.
Overall, within each morphology and treatment, the optimizer found controllers that behaved correctly, during training, under both the `move' and `stop' commands: Robots moved significantly more when commanded to do so than when commanded to `stop'
(green bars are significantly higher
than orange bars).

There was no
significant difference between control and experimental treatments, in any of the tested morphologies, in terms of the final displacement of optimized robots commanded to move
(pairs of green bars in each panel are of
equal height). 
This implies that training performance of the robots is not due to inherent properties of the word2vec embedding; rather, it is due to the evolutionary algorithm.

In both treatments and in five of the six morphologies (but not the quadruped) there is a significant difference between the displacement of robots given the training and testing `stop' commands 
(blue bars are usually higher than the
orange bar to their left).
Thus, the robots did not completely understand the meaning of the command `stop'.
This was somewhat expected given the distance between the variants of `stop' in the word2vec space (Table~II).
However, for four of the six morphologies, robots optimized with the experimental treatment moved less under the testing `stop' command than those optimized with the control treatment (red significance brackets in Fig.~\ref{fig:testError}).
Thus, morphology affects the grounding of the `stop' commands.

However, the training set is unbalanced|there are three
commands for `stop' and only 
two for `move'|so it is possible that robots are overfitting to the stop commands and thus display little motion during testing without learning the semantic meaning of the commands.

\begin{figure}
    \centering
    \includegraphics[width=0.8\linewidth]{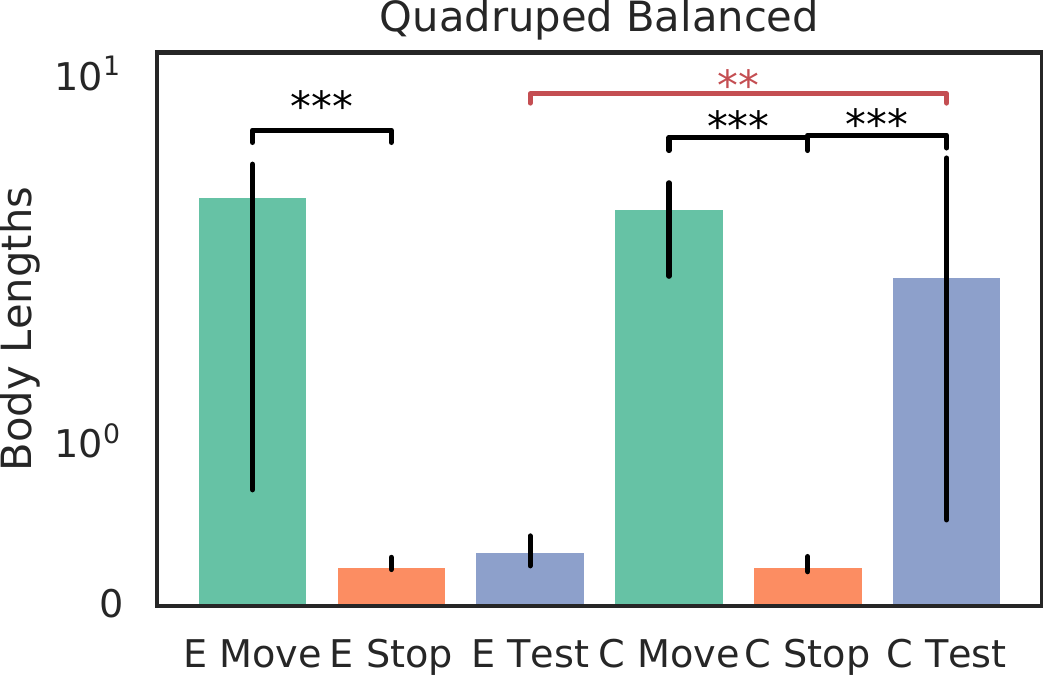}
    \caption{
    \textbf{Per-task balanced training sets.}
    To control for the possibility that generalization to the unheard ``stop'' command is actually due to overfitting behavior to the more prevalent ``stop'' commands, we retrained controllers for the quadruped from scratch using a balanced training set, consisting of two commands for each task: two ``stop'' commands, two ``forward'' commands, and two ``backward'' commands.
    Note that while this alternate training set is balanced per task, it is unbalanced in another way: there are four ``move'' commands but only one ``stop'' command.
    The significance levels under these alternative data were identical to those of the original (per-task unbalanced) regime (Fig.~\ref{fig:testError}c).
    Just as in Fig.~\ref{fig:testError}, the median displacement of the run champions is shown during training and under the unheard `stop' command (``Test''), for the experimental treatment (E; word2vec vectors) and the control treatment (C; randomly-permuted word2vec vectors).
    The seven pairwise comparisons made here are included in the 56 total comparison count used to correct p-values throughout the paper (details in \S2G).
    }
    \label{fig:quadBalanced}
    \vspace{-1em}
\end{figure}

To control for this, we retrained the quadruped from scratch on a new, balanced set of commands (Fig.~\ref{fig:quadBalanced}), where each task (`stop', `forward', and `backward') was trained using two commands,
yielding a training set size of six.
The two commands were chosen for each task such that the cosine similarity between them was similar to that of the stop synonyms previously used.
We chose to use `forward' and the misspelled `foward' for the `forward' task; 
and `backward' and `backwards' for the `backward' task. 
For the `stop' task, we removed the `halt' command, leaving `stop’,  `suspend' and `cease', one of which was randomly held-out at the beginning of each evolutionary trial for testing, and the others were used for training.
(Also, the reward function paired with the `stop' commands was changed to be inversely proportional to the robot's total movement, thus protecting against the perverse instantiation of oscillating around the origin.)

This alternate training set is balanced on a per-task level, however it is unbalanced on another level: 
there are four commands for `move' and only two for `stop'.

As seen in Fig. \ref{fig:quadBalanced},
under the per-task balanced training sets, the optimizer still found controllers that generated correct behavior during training.
There was no statistically significant difference between the control and experimental treatments in terms of training performance. 

Under the control treatment, the quadruped moved significantly more for the test `stop' command than the training `stop' commands. 
In fact they moved almost as much during the test `stop' command as in their training `move' commands.
Thus the control treatment, with the per-task balanced training set, yielded controllers that were overfit to the `move' commands.

Under the experimental treatment, however, there was no
significant difference between the movement of the quadruped under the training and test `stop' commands.
This suggests that, despite the higher prevalence of `move' commands in the per-task balanced training data, controllers learned latent structure of the embedding, and used this understanding to correctly generalize to the unheard `stop' synonym.

\section{Discussion}
\label{sectDiscussion}

One limitation of this work is that the command vector is loaded serially into the controller, prior to behavior, and is potentially overwritten by proprioception and touch sensor data during behavior.
Ideally, robots would be able to hear the commands throughout their evaluation periods, therefore allowing them to modulate their interpretation of the command based on action.
Further, this would allow dynamic communication 
with the robot.

One way to achieve this is with wider controller architectures: each element in the vector commands could have its own input synapse.
Then, the entire vector could influence action at each time step, and be updated during behavior.
Moreover, the controllers should also be deeper such that more complex (nonlinear and hierarchical) latent structure of the embedding can be learned.

Additionally, unlike other end-to-end methods that are mostly automated,
the method presented here still requires much
manual intervention: the investigator must create an objective function
for each grouping of action synonyms. To minimize such intervention,
in future work we wish to investigate
whether a small set of semantically and motorically orthogonal
objective functions can be created
that enables the robot to generalize not just to unheard synonyms of
training commands, but also to novel sequences of commands.

Finally, these experiments were conducted with simulated robots.
In future work, we would like to investigate how well this technique extends to physical robots. To this end, we have already 
performed some initial work to investigate how well the use of vector spaces for training robots works on physical systems.

\subsection{The physical robot.}

\begin{figure}[b]
    \vspace{-1em}
    \centering
    \includegraphics[width=0.85\linewidth]{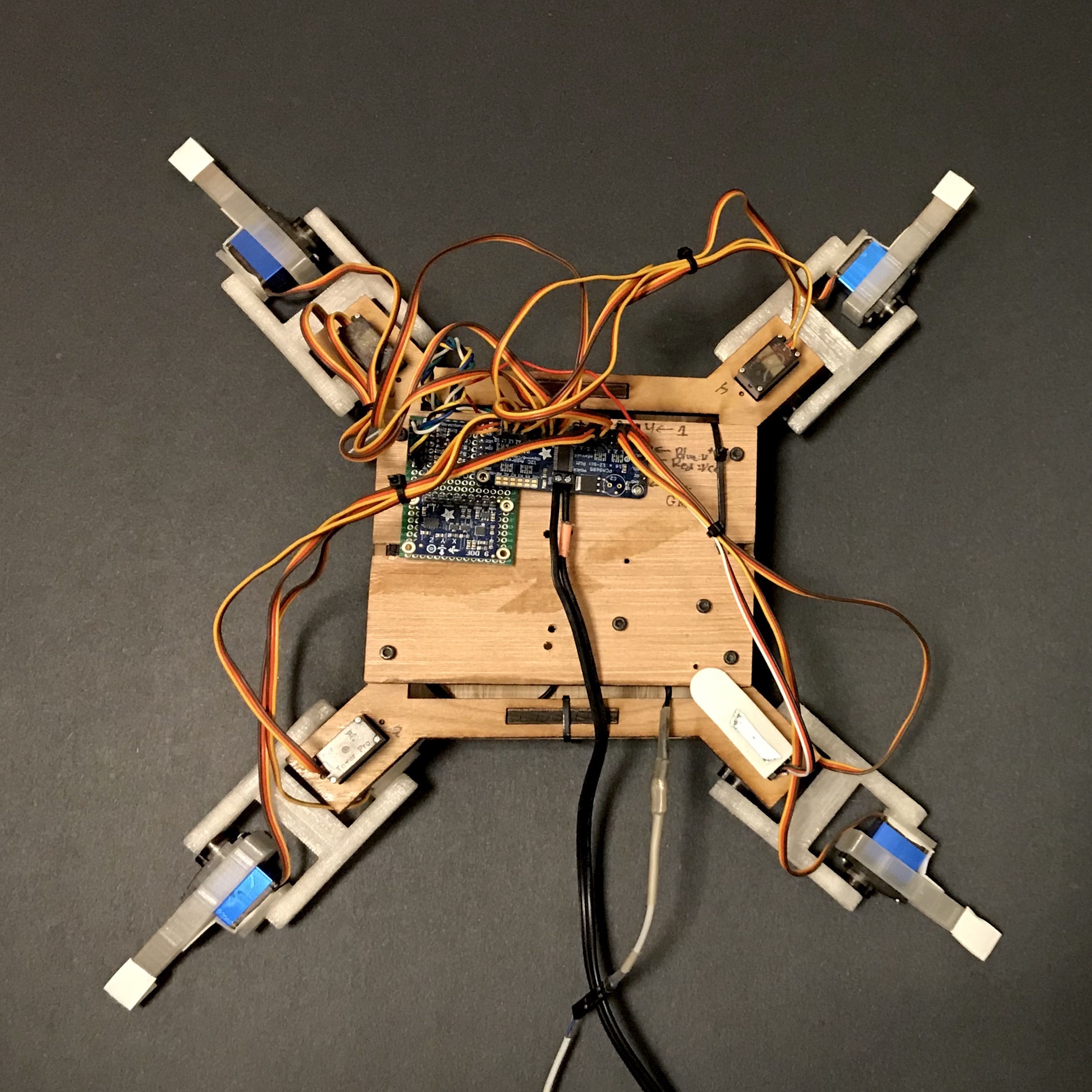}
    \caption{
        The physical 12 DOF quadruped.
    }
    \label{fig:physicalRobot2}
\end{figure}

\begin{figure*}
    \centering
    \includegraphics[trim={0 0 0 10em},clip,width=1\linewidth]{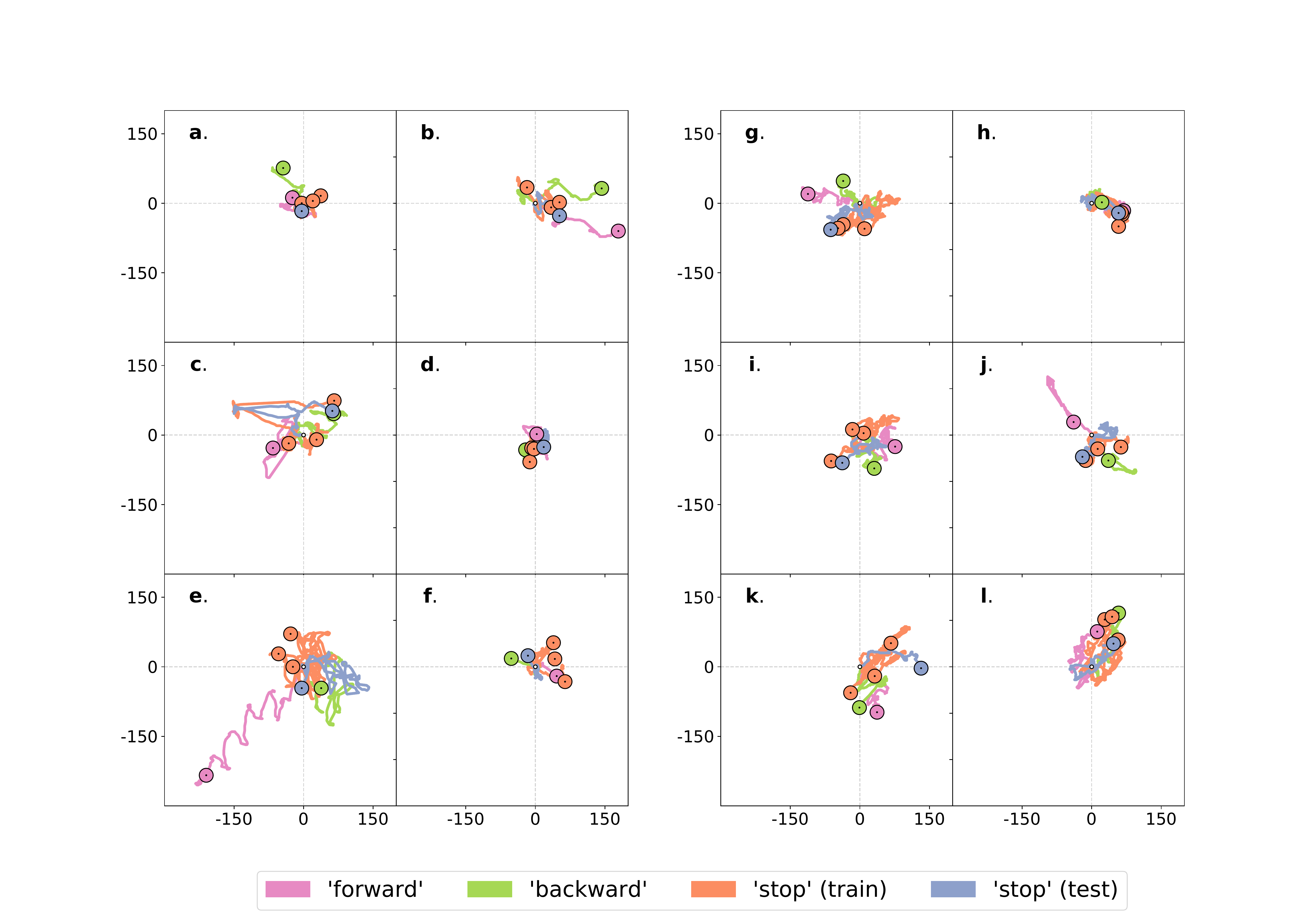}
    \vspace{-2em}
    \caption{\textbf{The trajectories of the physical robot.} 
        Twelve controllers were optimized in simulation and then transferred to the physical quadruped. 
        The movement of the physical robot is traced (in millimeters from the origin) for both the training and testing commands; final position at the end of the evaluation period of 3 seconds is denoted by square endpoints, colored differently for each command.
        The left panel (\textbf{a-f}) contains six run champions from the experimental treatment, and the right panel (\textbf{g-l}) contains six run champs from the control treatment. 
    }
    \label{fig:physicalTrajectories}
    \vspace{-1em}
\end{figure*}

Our physical robot system (Fig.~\ref{fig:physicalRobot2}) is a 12 DOF quadruped powered by a Raspberry Pi 3B+ and 12 14-gram Micro Servos. The main body is laser cut out of wood and holds the Raspberry Pi, an I2C PWM Driver, a 9DOF IMU, and a DC buck converter for power regulation. We provide power via an umbilical cord. Each of the four legs is constructed from 3D printed parts and contains three joints: a hip, a knee, and an ankle.

The robot is controlled by the on-board computer with programs written in Python.
The Python code interfaces with the robot sensors and motor driver over I2C to actuate the motors.
The sensor data can be added to the artificial recurrent neural network which is modeled by the Raspberry Pi.
An SSH connection over WiFi is used to perform maintenance, and configure and start the robot.

We created a new simulated quadruped without sensors and with a slightly modified morphology to more closely match the morphology of the physical robot.
Controllers were optimized in simulation under the same conditions as the original three robots.
We then transferred six optimized controllers from each treatment to the physical robot and recorded the motion using computer vision.
The movement patterns of these controllers are shown in Fig.~\ref{fig:physicalTrajectories}.

Overall, simulated behaviors did not transfer adequately to reality. However,
some of the controllers were able to exhibit movement denoting a minimally successful sim2real transfer.
For some of the controllers (e.g., Fig.~\ref{fig:physicalTrajectories}b,g,j), the physical robot moved in different ways for the `forward', `backward', and `stop' commands, thus exhibiting the rudiments of successful sim2real transfer.
Future work will more thoroughly investigate the use of vector spaces and existing sim2real methods \cite{bongard2006resilient,zhang2019vr,Hwangboeaau5872,kwiatkowski2019task} for training physical systems to ground language.

\section{Conclusions}
\label{sectConclusions}

In this work we have presented a method for inducing an
alignment between similarities among sensor data generated by
robot movements and the semantic similarities between the word2vec-encoded
commands that induced those actions. This method yields
control policies that cause robots to move appropriately
to previously-unheard natural language commands. Further,
we have found that this method can be facilitated or
frustrated by the particular mechanical structure of
the robot employed. In future work we plan to evolve
robot body plans, searching for those that make it even
easier to induce such alignments.
This work thus suggests not just that
relationships between action, human language, and embodiment
can be created in machines, 
but provides an empirical method for exploring and
strengthening these relationships to yield robots that
could be commanded by non-expert human handlers.

\section*{Source code}

\href{https://github.com/davidmatthews1uvm/2019-IROS}{\textcolor{blue}{\textbf{\texttt{github.com/davidmatthews1uvm/2019-IROS}}}}

\section*{Acknowledgements}

The authors would like to thank Eve Wight and Ryan Joseph for their help in creating the physical robot.
This work was supported by 
NSF award EFRI-1830870 and 
DARPA contract HR0011-18-2-0022.
Computation was provided by the 
Vermont Advanced Computing Core.

\bibliographystyle{IEEEtran}
\bibliography{root}

\end{document}